\documentclass{article}

\usepackage{arxiv}

\usepackage{cite}
\usepackage{amsmath,amssymb,amsfonts}
\usepackage{hyperref}
\usepackage{booktabs} 
\usepackage{algorithmic}
\usepackage{graphicx}
\usepackage{textcomp}
\usepackage{xcolor}
\usepackage{subcaption}     
\usepackage{multirow, tabularx, array}        
\usepackage{adjustbox}      

\def\BibTeX{{\rm B\kern-.05em{\sc i\kern-.025em b}\kern-.08em
    T\kern-.1667em\lower.7ex\hbox{E}\kern-.125emX}}

\title{Rethinking generalization of classifiers in separable classes scenarios and over-parameterized regimes}

\date{} 					

\author{ \href{https://orcid.org/0000-0002-8305-7138}{\includegraphics[scale=0.06]{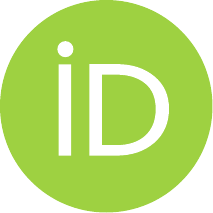}\hspace{1mm}Julius Martinetz} \\
	Machine Learning Group\\
	Technical University Berlin\\
	Berlin, Germany \\
	\texttt{j.martinetz@tu-berlin.de} \\
	\And
	\href{https://orcid.org/0000-0002-7039-5189}{\includegraphics[scale=0.06]{orcid.pdf}\hspace{1mm}Christoph Linse} \\
	Institute for Neuro- and Bioinformatics\\
	University of L{\"u}beck\\
	L{\"u}beck, Germany \\
	\texttt{c.linse@uni-luebeck.de} \\
    \And
	\href{https://orcid.org/0000-0002-4539-4475}{\includegraphics[scale=0.06]{orcid.pdf}\hspace{1mm}Thomas Martinetz} \\
	Institute for Neuro- and Bioinformatics\\
	University of L{\"u}beck\\
	L{\"u}beck, Germany \\
	\texttt{thomas.martinetz@uni-luebeck.de} \\
}

\renewcommand{\headeright}{}
\renewcommand{\undertitle}{}
\renewcommand{\shorttitle}{Rethinking generalization}

\hypersetup{
pdftitle={Rethinking generalization of classifiers in separable classes scenarios and over-parameterized regimes},
pdfauthor={Julius Martinetz, Christoph Linse, Thomas Martinetz},
pdfkeywords={Over-parameterized networks, generalization, small training data sets, Empirical Risk Minimization, flat minima},
}

\begin{document}
\maketitle

\begin{abstract}
We investigate the learning dynamics of classifiers in scenarios where classes are separable or classifiers are over-parameterized. In both cases, Empirical Risk Minimization (ERM) results in zero training error. However, there are many global minima with a training error of zero, some of which generalize well and some of which do not. We show that in separable classes scenarios the proportion of "bad" global minima diminishes exponentially with the number of training data $n$. Our analysis provides bounds and learning curves dependent solely on the density distribution of the true error for the given classifier function set, irrespective of the set's size or complexity (e.g., number of parameters). This observation may shed light on the unexpectedly good generalization of over-parameterized Neural Networks. For the over-parameterized scenario, we propose a model for the density distribution of the true error, yielding learning curves that align with experiments on MNIST and CIFAR-10.
\end{abstract}

\section{Introduction}

Deepening Neural Networks has brought about a significant advancement in various real-world object recognition tasks \cite{KrSuHi2012}. This trend extends to Neural Networks employed in Natural Language Processing \cite{Bert2018} and reinforcement learning \cite{DeepMindAtari2015}. Increasing the size of networks has proven effective in enhancing performance, and remarkably, even venturing into realms with much more network parameters than training samples, zero training error and memorizing the training samples, does not seem to compromise generalization \cite{ZhangBengio2017}\cite{linse_large_2023}. Belkin called this the "modern" interpolation regime' \cite{Belkin2019}.

This departure from conventional machine learning wisdom is noteworthy, as the generalization gap, representing the disparity between true error and training error, is no longer uniformly bounded within these over-parameterized regimes. This phenomenon becomes evident in scenarios involving random labeling, where over-parameterized classifiers achieve zero training error, yet the generalization error remains as high as random choice \cite{ZhangBengio2017}. 

In this paper, we explore classification scenarios characterized by zero training error. This is the case when the classifier is over-parameterized and also in situations where the classes are separable. A classifier receives inputs $x$ from an input distribution $P(x)$ and assigns a class label using a classifier function $h(x)$. The input $x$ has a true label $y$ with probability $P(y\vert x)$. For each input $x$, the classifier produces a loss $L(y, h(x))\in \{0,1\}$, $0$ if the classification is correct and $1$ if the classification is incorrect. The error of the classifier $h$ on the whole data distribution $P(x,y)=P(y\vert x)P(x)$ is given by the expected loss
\begin{equation*}
    E(h)=\int L(y,h(x))P(x,y) dxdy. 
\end{equation*}
$E(h)$ is also called true error, generalization error, true loss or risk. 

The classifier $h$ is selected from a function set $\cal H$, which, for instance, may be determined by the architecture of a Neural Network, and the specific $h$ is defined by the network parameters. Learning, in this context, involves the process of choosing an $h$ from the set $\cal H$
to minimize $E(h)$. This is commonly achieved through Empirical Risk Minimization (ERM). Let ${\cal S}=\{(x_1,y_1),... ,(x_n,y_n)\}$ be a so-called training set comprising $n$ input samples $x$ together with their labels $y$, independently drawn from $P(x,y)$. Empirical risk is defined as the average loss on the training set  
\begin{equation*}
    E_{{\cal S}}(h)=\frac{1}{n}\sum_{i=1}^n L(y_i,h(x_i)). 
\end{equation*}
$E_{{\cal S}}(h)$ is commonly referred to as the training error or empirical loss. Learning via ERM chooses an $h\in \cal H$ that minimizes $E_{{\cal S}}(h)$. However, $E(h)$ and $E_{{\cal S}}(h)$ may deviate. This deviation is called the generalization gap. 

When the generalization gap is small, achieving a low training error ensures a good solution with a small true error. Statistical learning theory, e.g., based on VC-dimension \cite{Vapnik1998} or Rademacher complexity \cite{BartlettM2002}, provides uniform bounds for the generalization gap. In this paper, we look at scenarios where we always have training error zero solutions in our function set $\cal H$. A special case are separable classes scenarios, where also solutions with true error zero exist in $\cal H$. In these cases, by using the VC-dimension it can be shown that the probability to have a "bad" classifier (i.e.\ with true error larger than $\varepsilon$) within the set ${\cal H}({\cal S})$ of zero training error classifiers is bounded by
\begin{equation}
\label{eq:boundVC}
\Pr\left\{\hbox{$\exists$ $h\in{\cal H}({\cal S})$ with $E(h)\geq\varepsilon$}\right\}\leq 2\left(\frac{2en}{d}\right)^{d} e^{-\frac{\varepsilon}{2}n}
\end{equation}
with $d$ as the VC-dimension of the function set $\cal H$ and $0\leq\varepsilon\leq 1$ (see, e.g., \cite{bookAnthonyBartlett}). This uniform bound is very general and valid for each classification problem given by $P(x,y)$. However, it comes with the cost that the number of training samples $n$ has to be significantly larger than $d$ for the bound to become non-vacuous. For example, even in the simplest case of a linear classifier in 2D with a VC-dimension of $d=3$, we need at least $n=410$ training samples for the bound to become non-vacuous for a generalization gap of $\varepsilon=0.1$. 

In the over-parameterized case, however, $d$ is much larger than $n$. Nevertheless, in practical applications learning takes place, and ERM provides solutions that generalize well, even in extremely over-parameterized regimes. Theoretical and empirical attempts have aimed to unravel this apparent "mystery". Some argue that this phenomenon can be attributed to the implicit regularization effects of stochastic gradient descent \cite{Neyshabur2017}, \cite{brutzkus2018}, \cite{Soudry2018}, \cite{Lyu2020}, \cite{advani2020}, \cite{Vardi2023}. Moreover, various novel algorithm-dependent uniform generalization bounds have been proposed to provide explanations \cite{NeyshaburColt2015}, \cite{Bartlett2017}, \cite{Golowich2018}, \cite{sanjeev2018}, \cite{LiLiang2018}, \cite{Kolter2019}, \cite{neyshabur2019}, \cite{Wei2019}, \cite{LiangPoggio2019}. Additional insight is given by the concept of the Neural Tangent Kernel (NTK) with its linearization at initialization, however, it requires extremely wide neural networks \cite{Jacot2018}, \cite{Chizat2018O}, \cite{AllenZhu2018LearningAG}, \cite{Arora2019OnEC}, \cite{Arora2019FineGrainedAO}, \cite{Sohl-Dickstein2020}, \cite{Min2021}. Finally, there is the well-known approach of algorithmic stability which can be applied \cite{hardt16}, \cite{mou18a}, \cite{lei2022}, \cite{Oneto2023}. However, there is empirical skepticism that these bounds are of any use in more general settings \cite{Jiang2020fantastic} or that the concept of uniform convergence bounds is inherently the right approach \cite{Nagarajan2019UniformCM}.

Inequality \eqref{eq:boundVC} bounds the probability that there is a "bad" classifier within the set of classifiers with zero training error. It does not care about how many there might be but wants to guarantee that there is not a single one left. In the following, we allow "bad" classifiers. As long as its fraction within the set of classifiers with zero training error is small, it should not be detrimental and the probability that ERM provides a good solution should still be high. This is much less than to enforce that the fraction of "bad" classifiers is zero. We show that this requires much less training data. Our perspective aligns with and supports the experimental findings and theoretical considerations in \cite{Goldstein2023loss} and \cite{martinetz2023highly}, providing an alternative explanation of the "mystery" of good generalization in over-parameterized regimes.   

\begin{figure*}[t]
	\centering
	\begin{subfigure}[t]{0.32\textwidth}
		\includegraphics[width=1.1\textwidth]{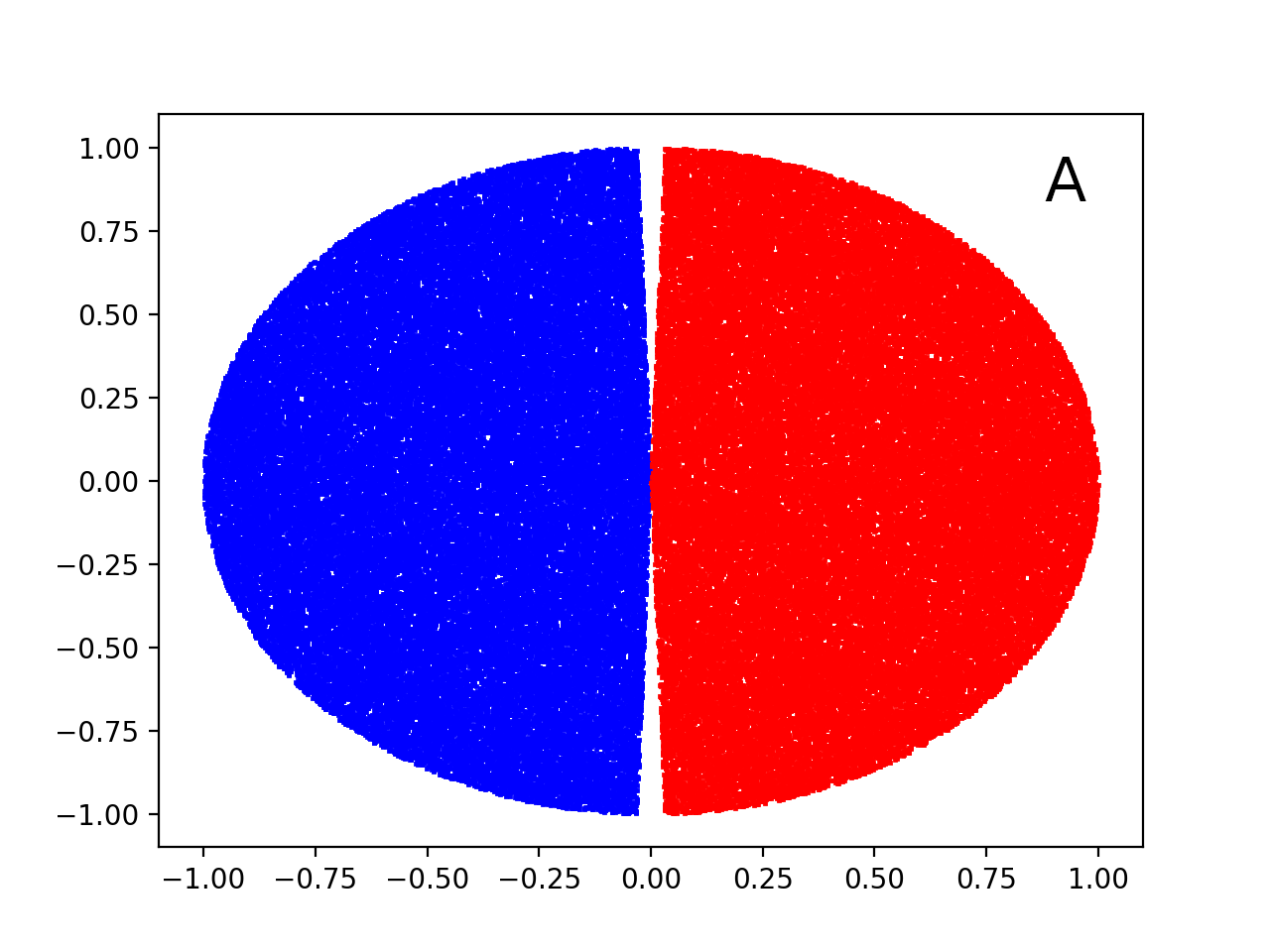}
	\end{subfigure}
    ~
	\begin{subfigure}[t]{0.32\textwidth}
		\includegraphics[width=1.1\textwidth]{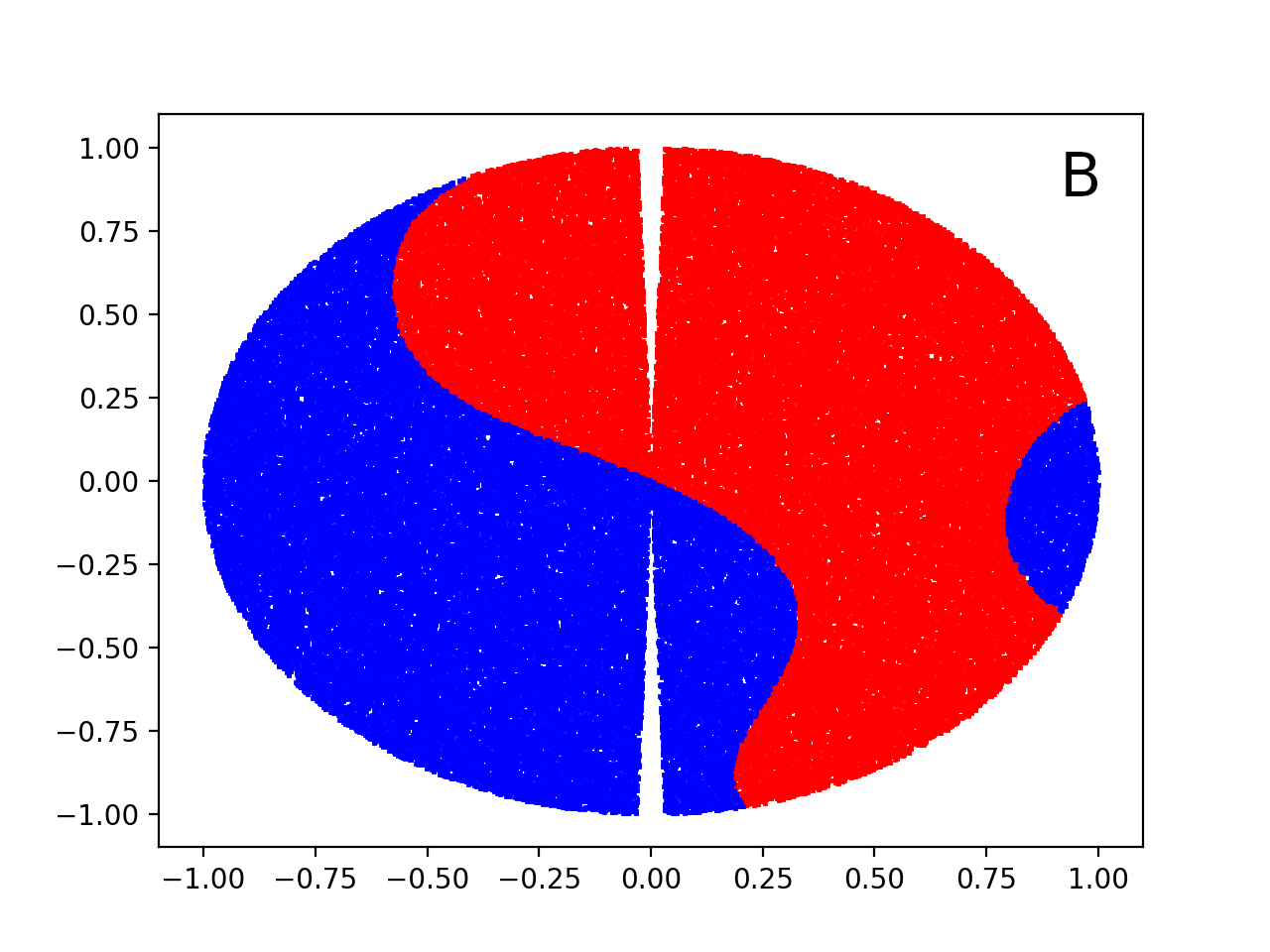}
	\end{subfigure}
    ~
	\begin{subfigure}[t]{0.32\textwidth}
		\includegraphics[width=1.1\textwidth]{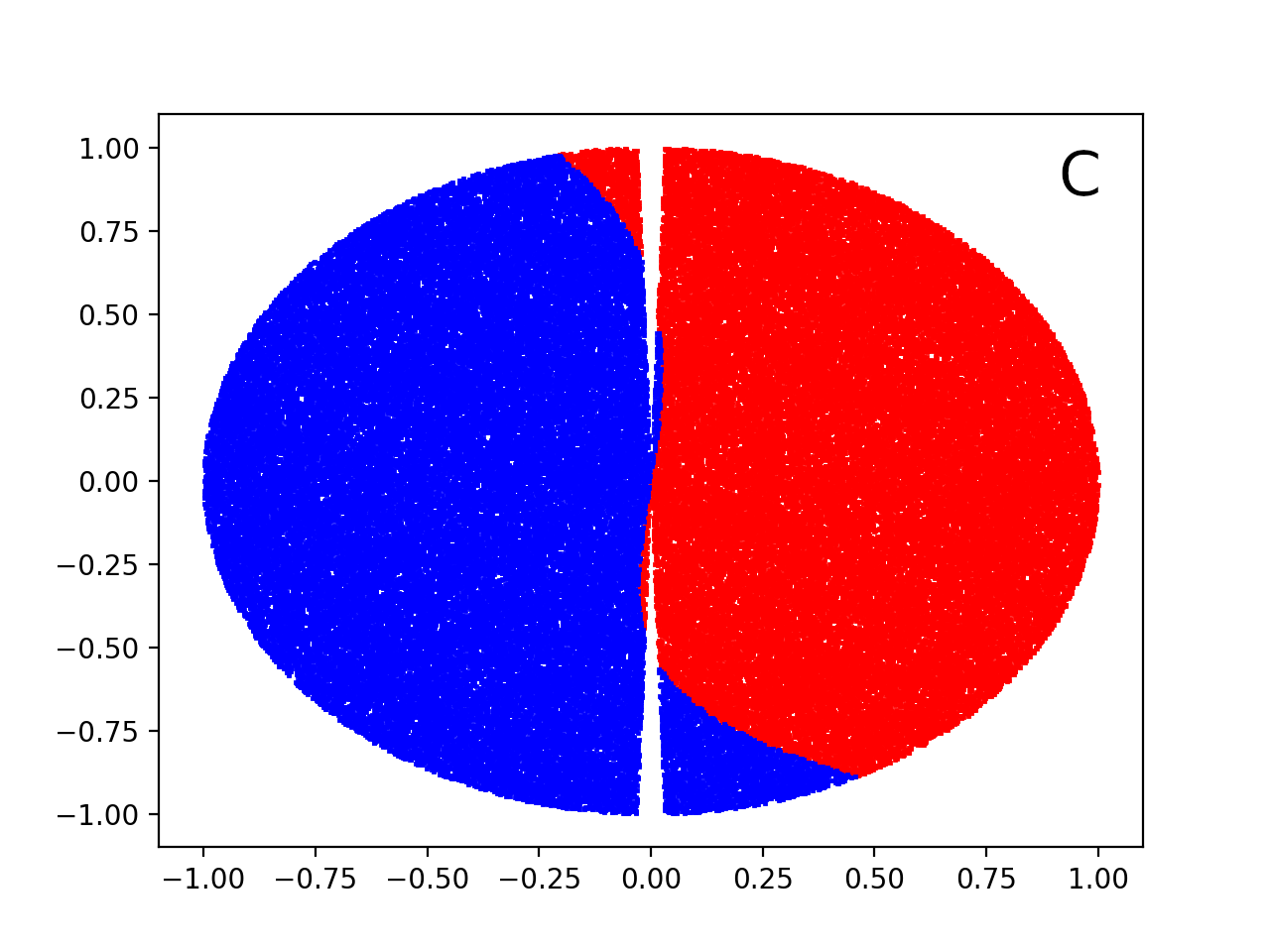}
	\end{subfigure}

	\caption{Classification problem in 2D with two (linearly) separable classes. The red and blue dots show 100000 random data points of both classes (A). The same data points classified with a polynomial classifier of degree 10 after training with 2 (B) and 20 (C) training samples.}
\label{fig:ToyProblem}
\end{figure*}

\section{Learning separable classes}

We start with the separable classes scenario. We assume that the classes do not overlap and, hence, are separable. This implies that to each $x$ of $P(x)$ always the same class $y(x)$ is assigned. We also assume that $\cal H$ is able to separate the classes, i.e., there is a subset ${\cal H}_0\subseteq \cal H$ such that each $h\in {\cal H}_0$ classifies any $x$ from $P(x)$ correctly (i.e., $L(y(x),h(x))=0$). Hence, $E(h)=0$ for each $h\in {\cal H}_0$. Since the training samples are drawn from $P(x)$, also $E_{{\cal S}}(h)=0$ for each $h\in {\cal H}_0$ and any training set ${\cal S}$. Hence, ERM\footnote{We assume that the ERM algorithm is able to find a global minimum of $E_{{\cal S}}(h)$. Convergence to global minima of ERM algorithms such as gradient descent or its variants is a topic on its own rights. However, in separable classes or over-parameterized settings convergence to zero training error is usually not a problem.} guarantees a perfect solution with $E(h)=0$, except that there exist any $h\in {\cal H}/{\cal H}_0$ which also have training error zero ($E_{{\cal S}}(h)=0$). In the following we take a look how many such "bad" $h$ exist in the solutions set. 

In order not to make it too technical, we consider function sets ${\cal H}$ that are discrete and finite. In fact, this assumption holds true whenever digital computers are utilized. Extending the basic concept to the continuous case is straightforward, but it becomes more technical and would distract from the basic idea. 
We work with the following subsets of ${\cal H}$:
\begin{eqnarray*}
{\cal H}_0&:& \hbox{set of all $h\in{\cal H}$ with $E(h)=0$}\\
{\cal H}_\varepsilon&:& \hbox{set of all $h\in{\cal H}$ with $E(h)\geq\varepsilon$}\quad (\varepsilon>0)\\
{\cal H}({\cal S})&:& \hbox{set of all $h\in{\cal H}$ with $E_{{\cal S}}(h)=0$}\\
{\cal H}_\varepsilon({\cal S})&:& \hbox{set of all $h\in{\cal H}_\varepsilon$ with $E_{{\cal S}}(h)=0$}.
\end{eqnarray*}
${\cal H}_0 \subseteq {\cal H}({\cal S})$ is valid, as well as ${\cal H}_\varepsilon({\cal S})\subseteq {\cal H}_\varepsilon$. 
For a given ${\cal S}$, any $h\in{\cal H}({\cal S})$ is a global minimum of the training error $E_{{\cal S}}(h)$. The fraction of "bad" global minima with $E(h)\geq \varepsilon$ is given by
\begin{eqnarray}
Frac\{E(h)\geq\varepsilon\vert {\cal S}\}&=& \frac{\vert {\cal H}_\varepsilon({\cal S})\vert}{\vert {\cal H}({\cal S})\vert} \label{eq:ProbX}\\
&\leq& \frac{\vert {\cal H}_\varepsilon({\cal S})\vert}{\vert {\cal H}_0\vert}.\nonumber
\end{eqnarray}
Taking the mean over all possible ${\cal S}$ yields a bound for the mean fraction of "bad" global minima 
\begin{equation}
\label{eq:bound}
    \left\langle Frac\{E(h)\geq\varepsilon\vert {\cal S}\}\right\rangle_{{\cal S}}\leq\frac{\langle\vert {\cal H}_\varepsilon({\cal S})\vert\rangle_{{\cal S}}}{\vert {\cal H}_0\vert}. 
\end{equation}
Now we derive an expression for the mean of $\vert {\cal H}_\varepsilon({\cal S})\vert$. For a given $h$, the loss $L(y,h(x))$ is a binomial random variable assuming the values $0$ or $1$ for training data drawn from $P(x,y)$. For $E_{{\cal S}}(h)=0$, $L(y_i,h(x_i))=0$ for each $(x_i,y_i)\in {\cal S}$. The probability of this arising with a random training set is $(1-E(h))^n$. With the indicator function $\mathbf 1_{{\cal H}_\varepsilon({\cal S})}(h)$, which is one for $h\in{\cal H}_\varepsilon({\cal S})$ and otherwise zero, we obtain
\begin{eqnarray}
    \langle\vert {\cal H}_\varepsilon({\cal S})\vert\rangle_{{\cal S}}&=&\left\langle\sum_{h\in {\cal H_\varepsilon}} \mathbf 1_{{\cal H}_\varepsilon({\cal S})}(h)\right\rangle_{{\cal S}} \nonumber\\
    &=&\sum_{h\in {\cal H_\varepsilon}} \left\langle\mathbf 1_{{\cal H}_\varepsilon({\cal S})}(h)\right\rangle_{{\cal S}} \nonumber\\
    &=&\sum_{h\in {\cal H_\varepsilon}} (1-E(h))^n \label{eq:sum}\\
    &\leq&\vert {\cal H}_\varepsilon\vert (1-\varepsilon)^n. \nonumber
\end{eqnarray}
The mean fraction of "bad" global minima within the set ${\cal H}({\cal S})$ of all global minima, i.e.\ those with generalization errors larger than $\varepsilon$, is then bounded by
\begin{eqnarray}
    \left\langle Frac\{E(h)\geq\varepsilon\vert {\cal S}\}\right\rangle_{{\cal S}}
    &\leq&\frac{\vert {\cal H}_\varepsilon\vert}{\vert {\cal H}_0\vert} (1-\varepsilon)^n \nonumber\\
    &\leq&\frac{\vert{\cal H}\vert-\vert{\cal H}_0\vert}{\vert {\cal H}_0\vert}(1-\varepsilon)^n.\label{eq:ourbound}\\
    &\leq& R e^{-\varepsilon n}\label{eq:ourboundexp}
\end{eqnarray}
with $R=(\vert{\cal H}\vert-\vert{\cal H}_0\vert)/\vert{\cal H}_0\vert$ and using $1-\varepsilon \leq e^{-\varepsilon}$. The bound goes to zero exponentially with the number of training samples $n$, and by doubling $n$ one obtains the same bound for half the $\varepsilon$. The prefactor $R$ is the size of the function set ${\cal H}$ relative to the size of its subset ${\cal H}_0$ containing the perfect classifiers. $0\leq R\leq\infty$ describes how well the function set fits the structure of the given problem. For $R=0$ we obtain a perfect fit where ${\cal H}={\cal H}_0$. Any global minimum thus provides a zero error (true and training). For $R\rightarrow\infty$ the fit becomes worse and we need more and more training data to reduce the mean fraction of "bad" global minima.


 An ERM algorithm for learning selects an $h\in{\cal H}({\cal S})$. Without any further knowledge, we expect the algorithm to choose any $h\in{\cal H}({\cal S})$ with the same probability (maximum entropy assumption). Then $Frac\{E(h)\geq\varepsilon\}$ gives the probability of ending up with a generalization error larger than $\varepsilon$. If, by any means such as, for instance, margin- or norm-based regularization, one succeeds in increasing the probability to end up in a "good" global minimum, the exponentially decreasing probability for "bad" solutions is even further reduced.  

\subsection{Experiments with separable classes}
To validate this bound we need (toy) problems where we know that the classes are (i) separable and (ii) separable by the function set that we use. A simple toy experiment is shown in Fig.~\ref{fig:ToyProblem}. The data distribution is homogeneous within a circle, except for two circle sections of $3.6^{\circ}$ each. The two classes do not overlap and are linearly separable.   

\begin{figure*}[t]
	\centering
	\begin{subfigure}[t]{0.48\textwidth}
		\includegraphics[width=0.99\textwidth]{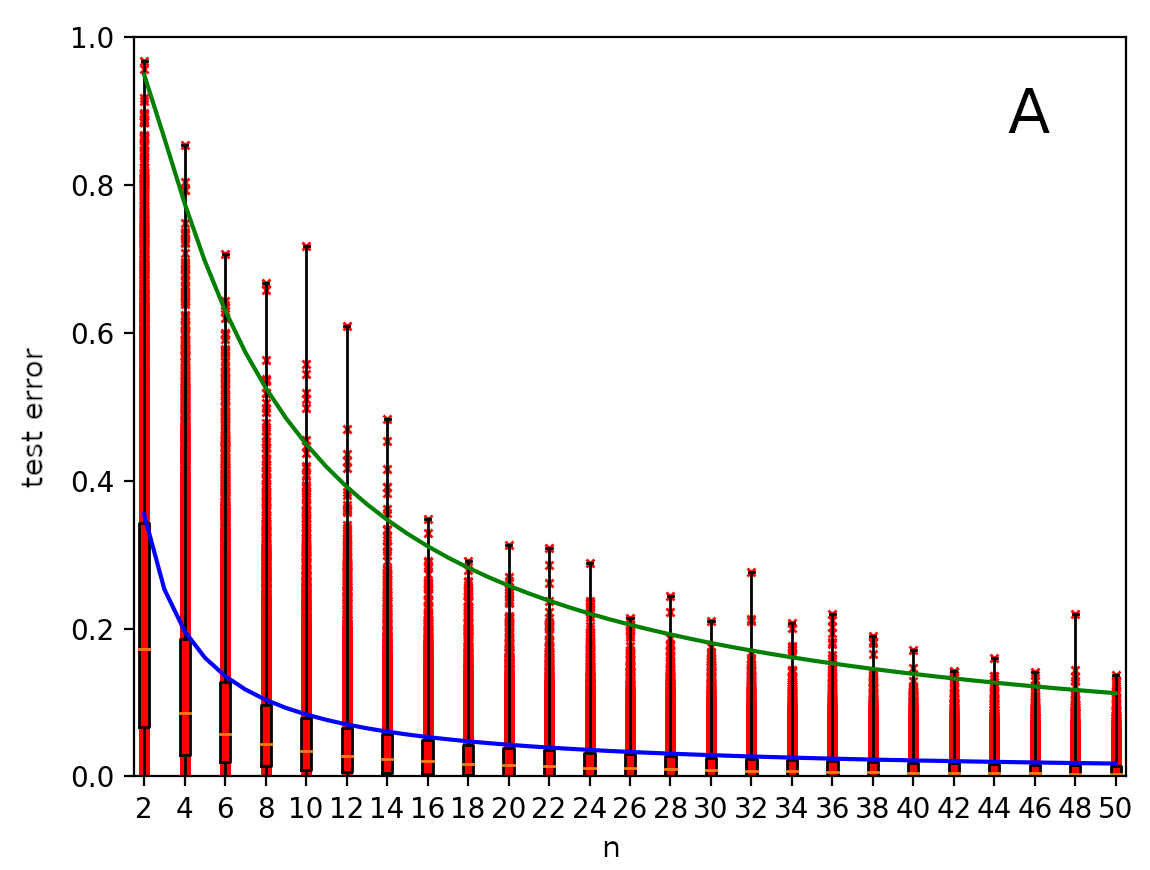}
	\end{subfigure}
    ~
	\begin{subfigure}[t]{0.48\textwidth}
		\includegraphics[width=0.99\textwidth]{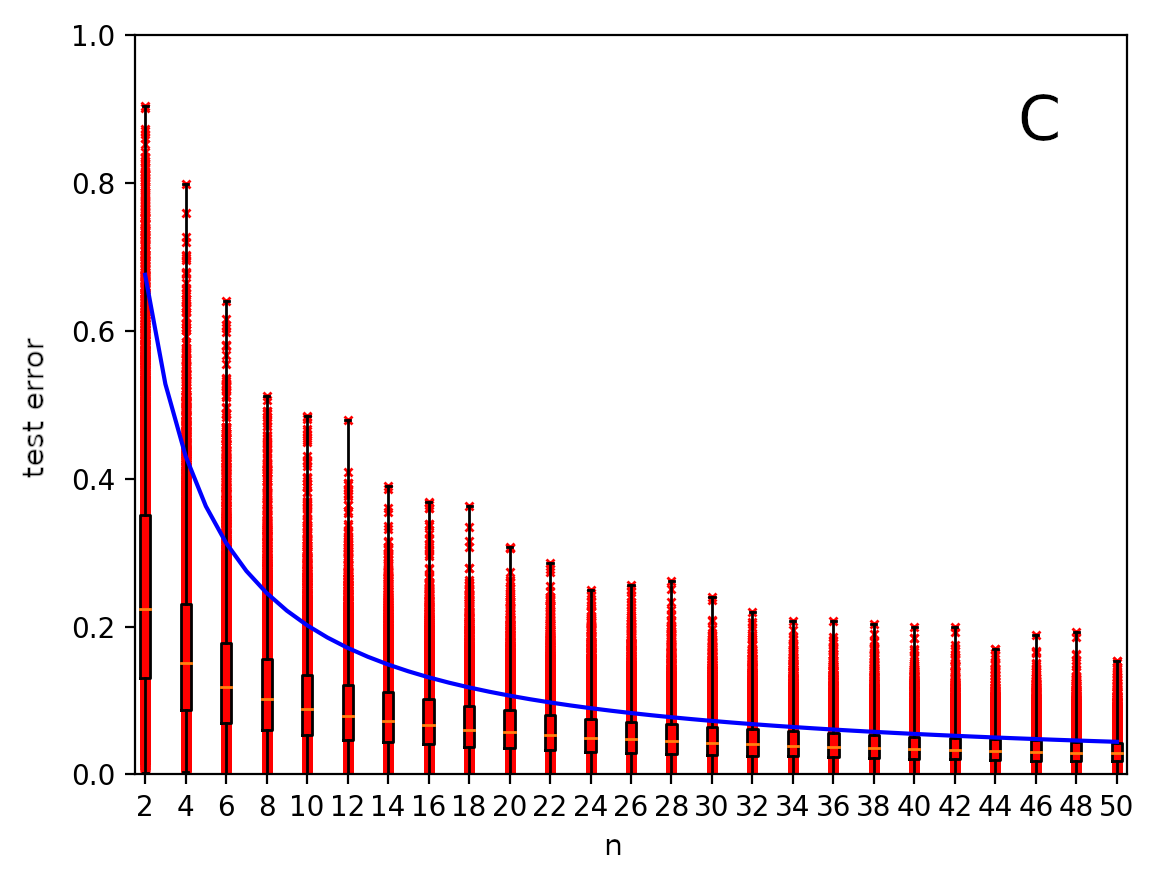}
	\end{subfigure}
	
	\begin{subfigure}[t]{0.48\textwidth}
		\includegraphics[width=0.99\textwidth]{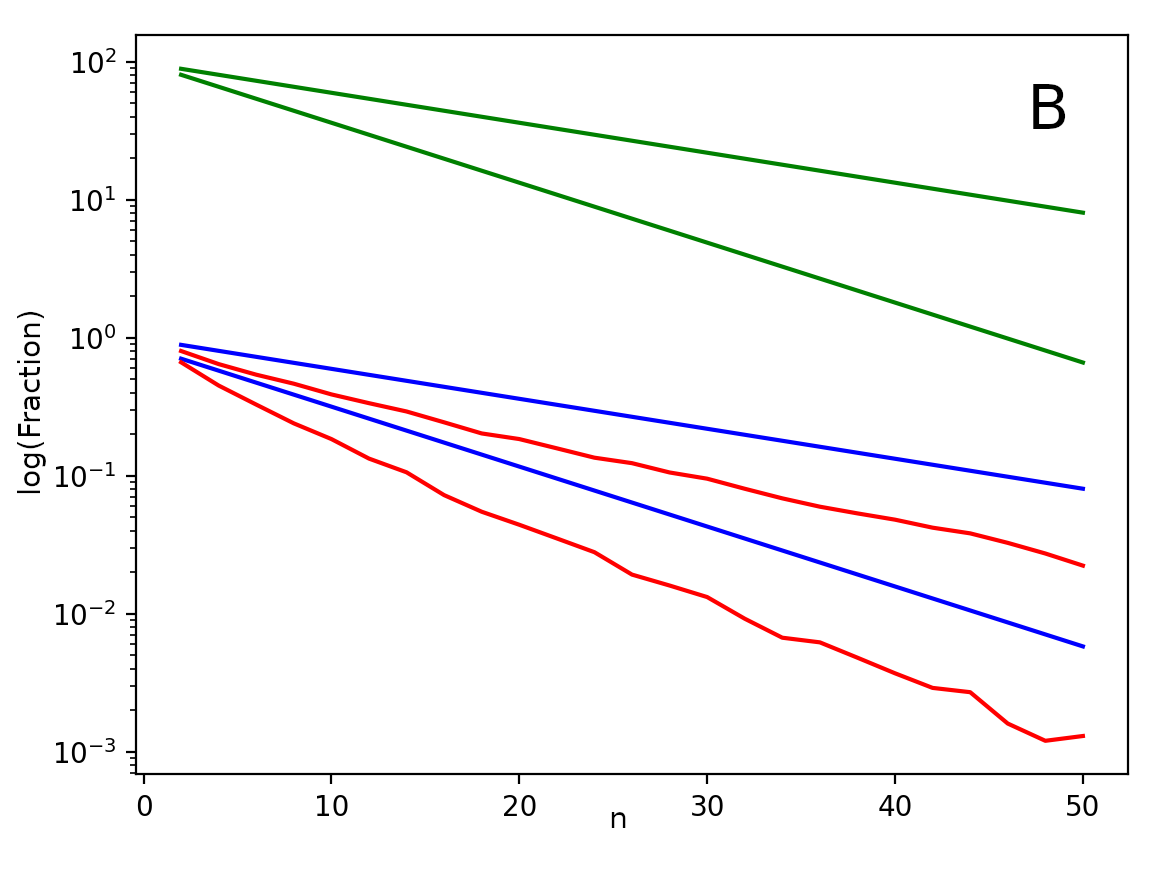}
	\end{subfigure}
    ~
	\begin{subfigure}[t]{0.48\textwidth}
		\includegraphics[width=0.99\textwidth]{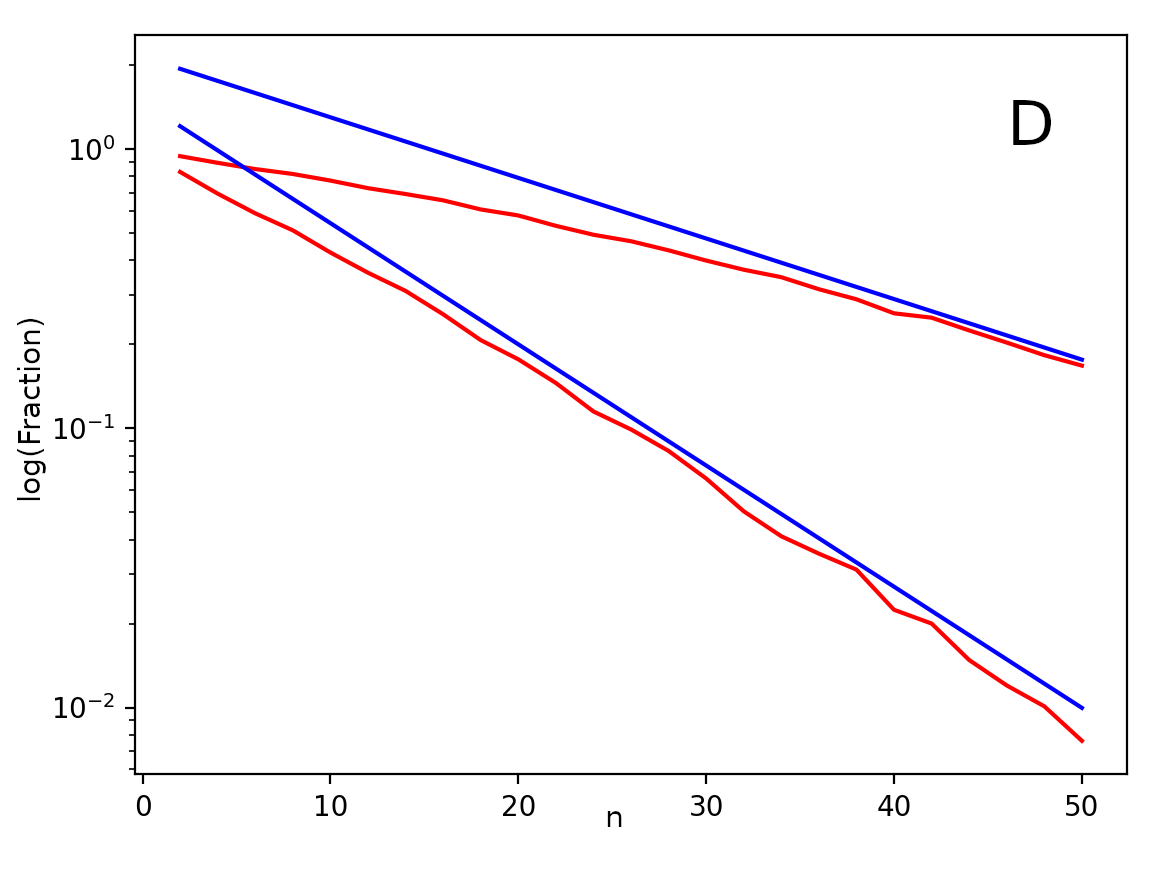}
	\end{subfigure}

    \caption{Distribution of the test errors (red crosses) for different numbers of training data on the classification problem shown in Figure~\ref{fig:ToyProblem}; for a linear classifier (A) and a polynomial classifier of degree 10 (C). The green line in A shows the 25\% bound given by inequality (\ref{eq:epsilonbound}) with $R=98$ and the blue line in A and C, respectively, with $R$ chosen such that the bound becomes tight. At the bottom, the red lines show the fractions of test errors exceeding $0.1$ and $0.05$, respectively, in a logarithmic plot, for the linear classifier (B) and for the polynomial classifier of degree 10 (D). The green lines in B are the bound (\ref{eq:ourboundexp}) with $R=98$ and the blue lines ind B and D with $R$ chosen such that the bound becomes tight, demonstrating the exponential decrease of the fraction of solutions with test errors exceeding a given $\varepsilon$.}
\label{fig:ResultsCircle}
\end{figure*}

We apply a linear classifier ${\bf w}^T{\bf x}$, with class $+1$ as output if ${\bf w}^T{\bf x}\geq 0$, and class $-1$ if ${\bf w}^T{\bf x}< 0$. As the ERM algorithm for learning, we use the perceptron algorithm, which guarantees convergence to zero training error if the classes are separable. The algorithm is simple: we start with a randomly-chosen normalized ${\bf w}_{t=0}$ and sweep through the training data. If a data point $x$ is misclassified, this data point is taken for an update of ${\bf w}$ according to 
\begin{equation*}
    {\bf w}_{t+1}={\bf w}_{t}+y{\bf x} 
\end{equation*}
with $y\in\{-1,+1\}$ as the true label of ${\bf x}$. Sweeping is repeated until all training data points are correctly classified. It is well established that this terminates after a finite number of training steps. 

If we take ${\bf w}\in \mathbb{R}^2$ and ${\bf x}=(x,y)$, with $x$ and $y$ as the 2D coordinates of the data points within the circle, the classifier is a straight line through the origin with a VC-dimension of two. According to the VC-based bound (\ref{eq:boundVC}), we need 363 data points for training to guarantee that in at most $25\%$ of all cases a generalization error worse than $0.1$ might occurs. Our bound (\ref{eq:ourbound}) tells us that 60 training data points are sufficient to have on average at most $25\%$ "bad" solutions within the ERM solution space with generalization errors worse than $0.1$. In order to calculate $R=(\vert{\cal H}\vert-\vert{\cal H}_0\vert)/\vert{\cal H}_0\vert$, we took $\vert{\cal H}\vert=360^{\circ}$ as the parameter volume comprising all possible directions of the classification line, and $\vert{\cal H}_0\vert=7.2^{\circ}$ comprising the solutions that separate the classes, i.e.\ $R=98$. In Fig.~\ref{fig:ResultsCircle}~A, we show the distribution of the test errors for different numbers of training data $n$. For each $n$, 10000 runs on randomly chosen data points were performed, and in each run it was trained until zero training error was reached. Each red cross is then the test error on 100000 randomly-chosen test data points. The box-plots show the medians and quartiles. As we can see, to end up with a generalization error worse than $0.1$ in $25\%$ of all cases, we even only need around 8 data points for training. By setting the right-hand side of inequality (\ref{eq:ourboundexp}) to $0.25$, we obtain the bounds for the $25\%$ quartile
\begin{eqnarray}
    \varepsilon_{25\%}&\leq& 1-\left(\frac{0.25}{R}\right)^{\frac{1}{n}}    \label{eq:epsilonbound}\\
    &\leq& \frac{\ln \frac{R}{0.25}}{n}    \label{eq:epsilonboundexp}   .
\end{eqnarray}
Bound \eqref{eq:epsilonbound} is the green line in Fig.~\ref{fig:ResultsCircle}~A. The bound is not tight, of course. The blue line shows bound (\ref{eq:epsilonbound}) for a $R$ that makes the bound tight. In Fig.~\ref{fig:ResultsCircle}~B, the green lines show a logarithmic plot of the right-hand side of bound \eqref{eq:ourboundexp} for $\varepsilon=0.1$ and $\varepsilon=0.05$, with slopes of $0.1$ and $0.05$. The red lines show in how many cases of the 10000 runs for each $n$ the test error surpasses $0.1$ and $0.05$, respectively, measuring the left-hand side of bound \eqref{eq:ourboundexp}. The blue lines depict the bounds (\ref{eq:ourboundexp}) for an appropriate $R$ that renders them tight, which clearly demonstrates the validity of the bounds.

In a second scenario we take ${\bf w}\in \mathbb{R}^{65}$ and ${\bf x}=(x,y,x^2,xy,y^2, ...,x^{10},x^9y,x^8y^2, ...,x^2y^8,xy^9,y^{10})\in\mathbb{R}^{65}$, i.e., a polynomial classifier of degree 10. It is still linear in its parameters ${\bf w}$. In this case the classifier is over-parameterized with its $65$ parameters and $n\leq 50$. According to the VC-based bound (\ref{eq:boundVC}), we should need at least 8592 data points for training to guarantee that in at most $25\%$ of all cases a generalization error worse than $0.1$ might occurs. In fact, as we can see in Fig.~\ref{fig:ResultsCircle}~C and D, only around 16 samples are sufficient to obtain a generalization error worse than $0.1$ in $25\%$ of all cases. This is not much more than for ${\bf w}\in \mathbb{R}^2$. A higher-dimensional parameter space does not necessarily imply a much larger $R$. Figure~\ref{fig:ResultsCircle}~D confirms very nicely the $\varepsilon n$ dependence in the exponent of bound \eqref{eq:ourboundexp}.

\section{Learning of over-parameterized classifiers}

Over-parameterized classifiers have a capacity with respect to the training dataset that leads to zero training error when employing an ERM algorithm. Hence, ${\cal H}({\cal S})$ is not empty and Equation~(\ref{eq:ProbX}) applies. Let $E_{min}=\min_{h\in{\cal H}} E(h)$ be the minimum error of the classifier function set ${\cal H}$ on the given classification problem. Averaged over all possible ${\cal S}$ analogues to Equation~(\ref{eq:bound}), we obtain 
\begin{eqnarray}
\left\langle Frac\{E(h)\geq\varepsilon\vert {\cal S}\}\right\rangle_{{\cal S}}&=&\left\langle \frac{\vert {\cal H}_\varepsilon({\cal S})\vert}{\vert {\cal H}({\cal S})\vert}\right\rangle_{{\cal S}} \nonumber\\
&\approx& \frac{\langle\vert {\cal H}_\varepsilon({\cal S})\vert\rangle_{{\cal S}}}{\langle\vert {\cal H}({\cal S})\vert\rangle_{{\cal S}}}. \label{eq:ProductAverages}
\end{eqnarray}
The last step results from a first order Taylor expansion \cite{KendallStatistics2006} and is valid, if, e.g., $\vert {\cal H}_\varepsilon({\cal S})\vert$ and $\vert {\cal H}({\cal S})\vert$ are concentrated around their mean. This and further scenarios for Equation~\eqref{eq:ProductAverages} being valid is discussed in the Appendix and in \cite{martinetz2023highly}. It is further supported empirically by the results of our experiments. Assuming that this step is applicable, we can use the derivation of Equation~(\ref{eq:sum}) and obtain 
\begin{eqnarray}
\left\langle Frac\{E(h)\geq\varepsilon\vert {\cal S}\}\right\rangle_{{\cal S}}&=&\frac{\sum_{h\in {\cal H_\varepsilon}} (1-E(h))^n}{\sum_{h\in {\cal H}} (1-E(h))^n}\nonumber \\
&=&\frac{\int_{\varepsilon}^1  (1-E)^n D(E)\,dE}{\int_{0}^1 (1-E)^n D(E)\,dE }\nonumber\\
&=&\int_{\varepsilon}^1  Q_n(E)\, dE. \nonumber
\end{eqnarray}
$D(E)$ is the density of classifiers (DOC) at true error $E$, with $D(E)dE$ counting the number of classifiers functions $h\in {\cal H}$ with $E\leq E(h) \leq E+dE$ (analogous to the density of states (DOS) in solid-state physics). Accordingly, 
\begin{equation}
    Q_n(E)=\frac{(1-E)^n D(E)}{\int_{0}^1 (1-E)^n D(E)\,dE }
    \label{eq:Q_n}
\end{equation}
is the average (normalized) density of global minima at true error $E$, given $n$ training data points. $Q_n(E)$ decays to zero exponentially with increasing $n$ for each $E>E_{min}$ (see Appendix).

The true error averaged over all global minima of all ${\cal S}$ is then given by
\begin{eqnarray}
\left\langle E_n\right\rangle &=& \int_{0}^1 E\, Q_n(E)\,dE \\
&=&\frac{\int_{0}^1 E(1-E)^n D(E)\,dE }{\int_{0}^1 (1-E)^n D(E)\,dE }\label{eq:error}.
\end{eqnarray}

If the ERM algorithm for training chooses any global minimum with equal probability (maximum entropy assumption), this gives us the learning curve with respect to the number of training data $n$. 
The true error we can expect after training depends on $D(E)$, the density of classifiers (DOC). Due to the normalization in Equation~(\ref{eq:error}), it is the shape of $D(E)$ and not its absolute density values that determines the learning curve. $D(E)$ depends on the ${\cal H}$ that we choose and the given classification problem. However, since its shape is relevant, learning does not necessarily depend on the size, capacity or number of parameters of ${\cal H}$. Increasing the capacity may keep the shape invariant. 

As a side remark: for small values of $E$ we have $1-E\approx e^{-E}$. For large $n$ only small $E$ values contribute to the integrals in Equation~(\ref{eq:error}), and thus we can write
\begin{eqnarray*}
    \left\langle E_n\right\rangle &=&\frac{\int_{0}^1 E\, e^{-nE} D(E)\,dE }{\int_{0}^1 e^{-nE} D(E)\,dE }\\
        &=&-\frac{\partial}{\partial n} \log Z_n
\end{eqnarray*}
with $Z_n=\int_{0}^1 e^{-nE} D(E)\,dE$ as the partition function of the system. This corresponds to a physical system, with $h\in {\cal H}$ as its states, $E(h)$ as the energy of state $h$, $D(E)$ as the density of states (DOS) at energy $E$, and $e^{-nE}$ as the Boltzmann factor with $n$ as the inverse temperature. With increasing $n$, the temperature goes to zero and the system will mainly be found in its ground state with energy (error) $E_{min}$.

\subsection{A model for the density of classifiers (DOC)}

We have $D:\left[0,1\right]\rightarrow \mathbb{R}_0^+$, and $D(E)$ starts to become non-zero for $E\geq E_{min}$ with increasing $E$. In the non-separable case, we have $E_{min}>0$, and in a typical setting we expect $D(E)$ to increase monotonically with $E$ up to a maximum value at $E_0$, which corresponds to random choice. Then $D(E)$ decreases monotonically to become zero again for $E\geq E_{max}$. In binary classification problems and if for each $h\in {\cal H}$ there is also an $\bar h\in {\cal H}$ with $\bar h$ assigning exactly the opposite class to each input, which is typically the case and simply achieved by switching signs appropriately, $D(E)$ is symmetric around $E=1/2$ and $E_{max}=1-E_{min}$. Another example is the case of random labelling as experimentally studied in \cite{ZhangBengio2017} on CIFAR-10 with 10 classes. Then for each $h$ the true error is $0.9$ corresponding to random choice. Hence, $D(E)=\delta(E-0.9)$ with $\delta(.)$ as the Dirac-function, and with Equation~(\ref{eq:error}), we obtain $\left\langle E_n\right\rangle=0.9$ for each $n$. As expected, there is no generalization from training error zero also for large $n$.    

\begin{figure}[t]
	\centering
	\includegraphics[width=0.99\linewidth]{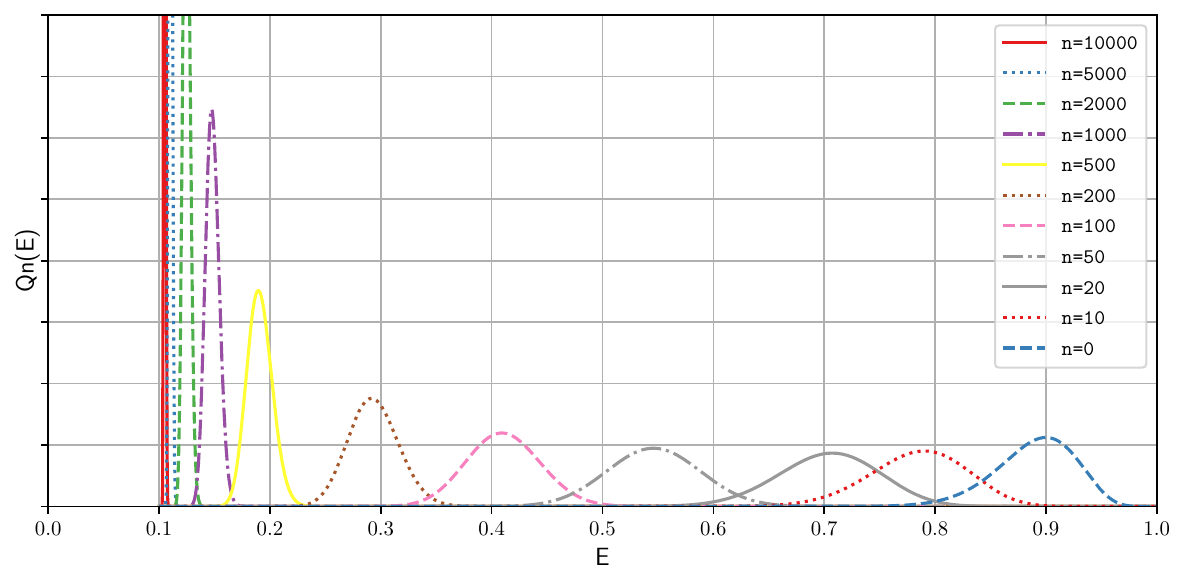}
	\caption{$Q_n(E)$ with $E_{\textrm{min}} = 0.1$, $\alpha=57$ and $\beta=8$. For $n=0$ we obtain $D(E)$ with its maximum at $E=0.9$. For large $n$ we had to cut the tips of the curves.}
	\label{fig:Qn_e}
\end{figure}

A good model of $D(E)$ results in a good model for the learning curve. For our experiments in the next section we model the monotonic increase of $D(E)$ as $(E-E_{min})^{\alpha-1}$, with $\alpha$ as a kind of intrinsic dimensionality of ${\cal H}$ with respect to $E$, and $E-E_{min}$ as the distance (radius) from the origin at the minimum. Accordingly, we model the decrease as $(E_{max}-E)^{\beta-1}$. This leads to 
\begin{equation*}
    D(E)=(E-E_{min})^{\alpha-1} (E_{max}-E)^{\beta-1} 
\end{equation*}
as a model for $D(E)$. In Fig.~\ref{fig:Qn_e}, the shape of $D(E)=Q_0(E)$ is shown for $E_{min}=0.1$ and $\alpha=57$, $\beta=8$, two values which sum up to $65$ as later observed in an experiment with ResNet on MNIST. Its maximum lies at $E=0.9$ (random choice). Figure~\ref{fig:Qn_e} also shows $Q_n(E)$ for different values of $n$. For comparison, the area under each curve is normalized. The position of their maximum shifts to smaller $E$ values with increasing $n$, and, at the same time, their width becomes small. For $n=10000$, the height of $Q_n(E)$ is ten times that of $n=1000$ (we had to cut the heights in Fig.~\ref{fig:Qn_e}), and its width is about one tenth. $Q_n(E)$ as the average distribution of the true error within the set of global minima shows, that almost all global minima have the same true (i.e.\ generalization) error. It is very unlikely to end up with "bad" generalization.

In our experiments in the next section we work with MNIST \cite{lecun-mnisthandwrittendigit-2010} and CIFAR-10 \cite{cifar10}, each having 10 classes. In this case, random choice leads to an error of $0.9$ and, hence,  $0.9\leq E_{max}\leq 1$. Therefore, in many classes scenarios we may take $E_{max}\approx 1$, which allows to solve Equation~(\ref{eq:error}) in closed form (e.g.\ take a computer algebra system) and gives the learning curve 
\begin{eqnarray}
    \left\langle E_n\right\rangle &=& \frac{E_{min}(\beta+n)+\alpha}{\alpha+\beta+n}\nonumber\\
    &=&E_{min}+\frac{E_0-E_{min}}{1+n/\eta}\label{eq:modelerror}
\end{eqnarray}
with $\eta=\alpha+\beta$ and $E_0=(\alpha+\beta E_{min})/(\alpha+\beta)$ as the error for $n=0$. Typically, $E_0$ corresponds to random choice. The true error that we can expect after training goes to the minimal achievable value with $1/n$ for large $n$. "Large" is measured with respect to the scaling factor $\eta$, determined by the "dimensionalities" $\alpha$ and $\beta$. According to this learning curve, a good choice for ${\cal H}$ is one with a small $E_{min}$ and a small $\eta$. In the following, we will show in our experiments how well the learning curve based on our model assumptions on $D(E)$ fits in typical scenarios.

\subsection{Experiments with over-parameterized classifiers}

We start with Convolutional Neural Networks (CNNs) \cite{CNN1989} applied to the MNIST dataset \cite{lecun-mnisthandwrittendigit-2010}. MNIST comprises 70000 images of hand-written digits (0,1, ..., 9), 60000 for training and 10000 for testing. Each image has a size of 28x28 pixels. As CNNs, we choose ResNet018 and ResNet101 \cite{HeZRS15}, two widely-used architectures with 11.2 and 42.5 million parameters, respectively. Obviously, both are highly over-parameterized. 

Both networks were trained for various $n$ (10, 20, 50, 100, 200, 500, ..., 15000, ..., 60000), always such that a global minimum of zero training error is reached at the end. For each $n$, 50 runs were performed. Details about the architectures and the training can be found in the Appendix. In Fig.~\ref{fig:modelerror_mnist}, the mean test error and its standard deviation over the 50 runs for each $n$ is shown for ResNet018 and ResNet101. The red curves show Equation~\eqref{eq:modelerror} with $E_{min}$ and $\eta$ adapted to the mean test errors. Both networks converge quickly to very small test errors, despite extreme over-parameterization and memorizing the training data. Therefore, we also show a double logarithmic plot.  Table~\ref{tab:modelerror_parameters} shows the values of $E_{min}$ and $\eta$. As expected, ResNet101 has a smaller $E_{min}$ as ResNet018. However, $\eta$ is also smaller for ResNet101. Hence, ResNet101 converges faster to its $E_{min}$ than ResNet018. For a fixed $n$, ResNet101 generalizes better than ResNet018 in this over-parameterized regime, despite there being four times the number of parameters. This is in accordance with the observations in \cite{Belkin2019}. ResNet101 easily achieves state-of-the-art results on MNIST without any data augmentation. Note, that the double logarithmic plot shows the deviation from the respective $E_{min}$.

\begin{figure}[t]
	\centering
	\begin{subfigure}[t]{0.48\linewidth}
		\includegraphics[width=\linewidth]{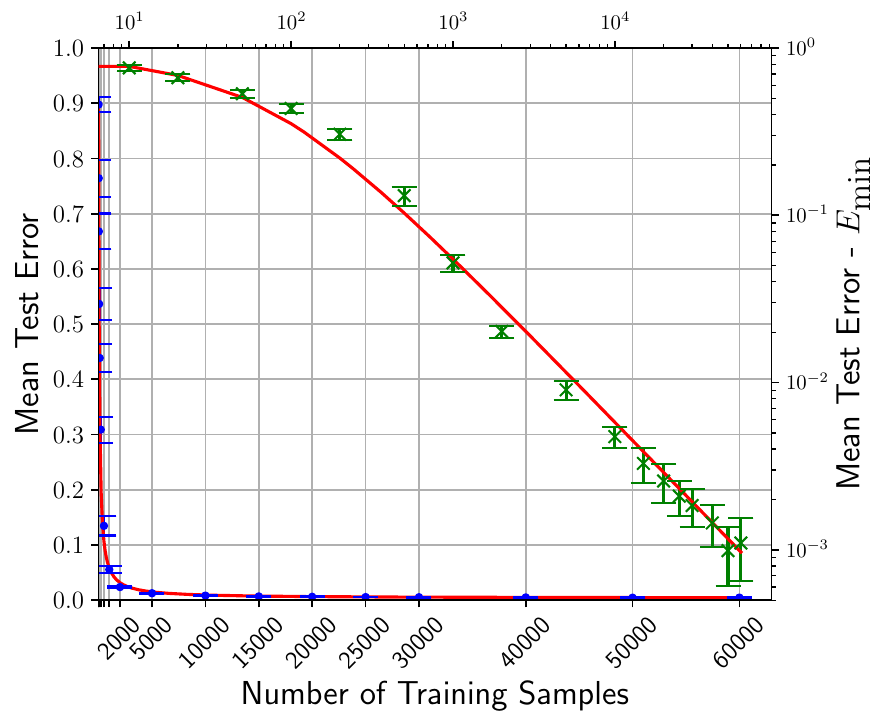}
	\end{subfigure}
	~
	\begin{subfigure}[t]{0.48\linewidth}
		\includegraphics[width=\linewidth]{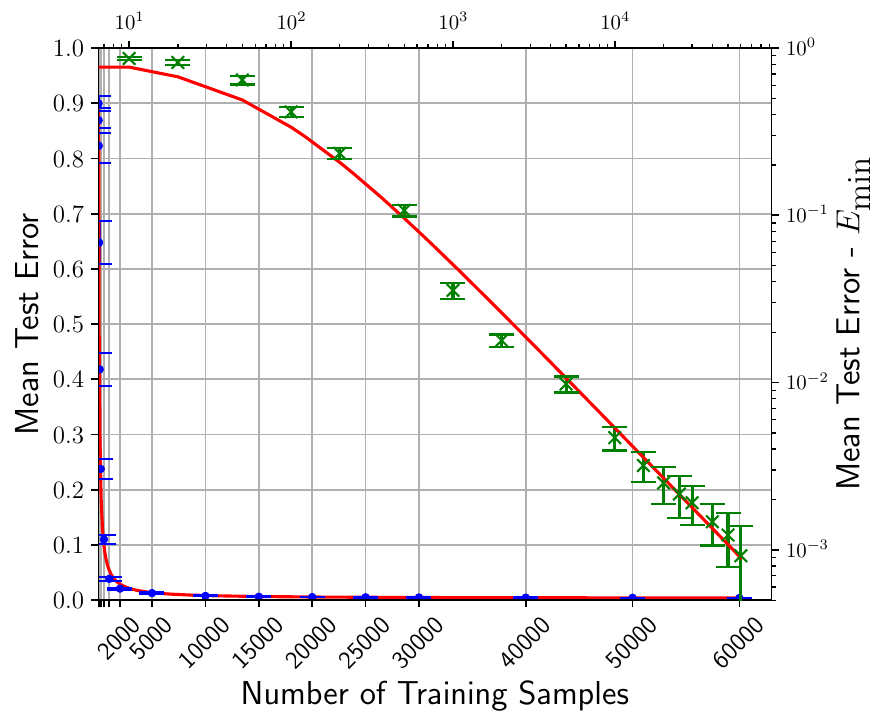}
	\end{subfigure}
	
	\caption{Mean test errors and their standard deviations of ResNet018 (top) and ResNet101 (bottom) on MNIST for different numbers of trainings samples $n$. The green crosses show their double logarithmic values as deviation from $E_{min}$. The red curves show fits of Eq.~\eqref{eq:modelerror} (parameters given in Table~\ref{tab:modelerror_parameters}).}
	\label{fig:modelerror_mnist} 
\end{figure}

\begin{table}[htb]
    \centering
    \begin{tabular}{l|l|l|lll}
    Dataset                   		& Architecture	& Parameters	& $E_0$		& $E_{\textrm{min}}$	& $\nu$	\\ \hline
    \multirow{2}{*}{MNIST}    		& ResNet018  	& 11.2M      	& 0.9    	& 0.0035  				& 65 	\\
									& ResNet101  	& 42.5M      	& 0.9    	& 0.003  				& 60 	\\ \hline
    \multirow{4}{*}{CIFAR-10} 		& MLP3       	& 12.8M      	& 0.9  		& 0.363  				& 4700 	\\
									& MLP8       	& 40,5M      	& 0.9  		& 0.360  				& 6000 	\\
                              		& ResNet018     & 11.2M      	& 0.9  		& 0.095  				& 4000 	\\
                              		& ResNet101     & 42.5M      	& 0.9  		& 0.067  				& 3700 	\\\\
    \end{tabular}
    \caption{Values of the parameters in Equation~\eqref{eq:modelerror} for specific architectures and datasets.}
	\label{tab:modelerror_parameters}
\end{table}

\begin{figure}[t]
	\centering
	\begin{subfigure}[t]{0.48\linewidth}
		\includegraphics[width=\linewidth]{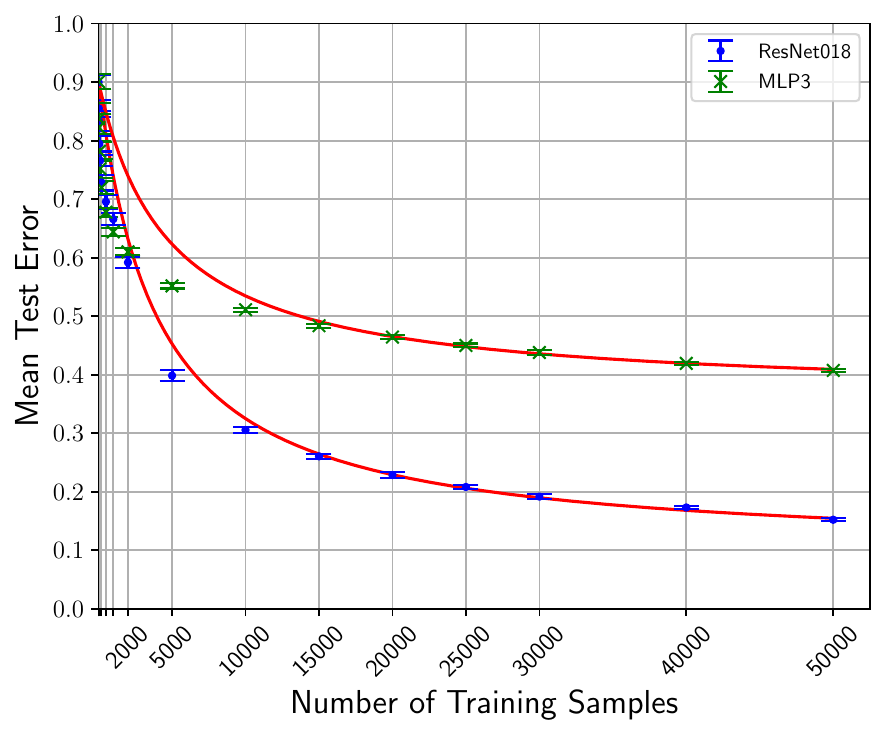}
	\end{subfigure}
	~
	\begin{subfigure}[t]{0.48\linewidth}
		\includegraphics[width=\linewidth]{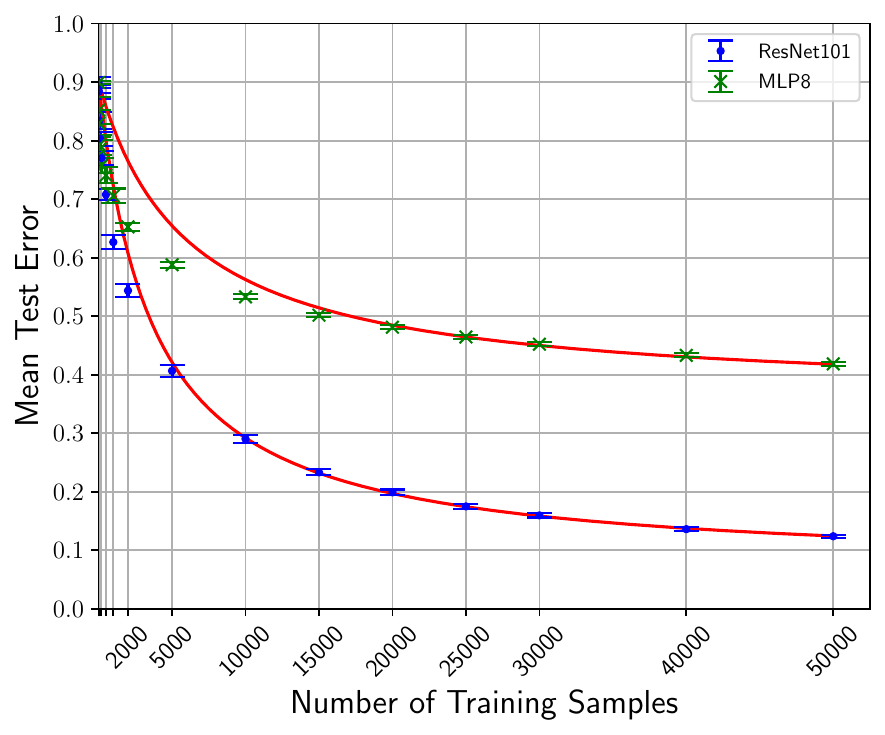}
	\end{subfigure}
	
	\caption{Mean test errors and their standard deviations of ResNet018 and MLP3 (top) and ResNet101 and MLP8 (bottom) on Cifar-10 for different numbers of trainings samples $n$. The red curves show fits of Eq.~\eqref{eq:modelerror} (parameters given in Table~\ref{tab:modelerror_parameters}).}
	\label{fig:modelerror_cifar10} 
\end{figure}

In a second step, we apply ResNet018 and ResNet101 to CIFAR-10 \cite{cifar10}. CIFAR-10 comprises 60000 color images of ten different object classes (cars, cats, ...), 50000 for training and 10000 for testing. Each image has a size of 32x32 pixels. For the purpose of comparison, this time we also train fully connected Multi-Layer-Perceptrons (MLPs) \cite{Rumelhart1986} with about the same number of parameters. For comparison with ResNet018, an MLP with 3 layers (MLP3) was taken, and for ResNet101 an MLP with 8 layers (MLP8). In Fig.~\ref{fig:modelerror_cifar10}, the mean test errors and their standard deviations are shown for different numbers $n$ of training samples. As for MNIST, 50 runs for each $n$ were performed, with training errors always being zero after training. First of all, the $E_{min}$ of the MLPs is much larger than of the ResNets. As can be seen in Table~\ref{tab:modelerror_parameters}, again for ResNet101 not only is $E_{min}$ smaller than for ResNet018, but also $\eta$. In contrast to the MLPs, where the larger network also has a larger $\eta$, i.e., a slower convergence. Details can be found in the Appendix.      

The red curves, showing our model, fit well but can still deviate from the experimental data, in particular for small $n$ and for the MLPs. The real test errors are smaller than predicted for
small $n$ on CIFAR-10. Of course, the real DOCs may deviate from our model, e.g., the intrinsic dimensionalities $\alpha$ and $\beta$ may not be constant and depend on $E$ or even be fractal. Also, the approximation $E_{max}\approx 1$ affects in particular small $n$ and leads to values that are too high. And there is approximation step \eqref{eq:ProductAverages}, which one might expect to be less accurate for small $n$. There is plenty of room for further insights and improvements.   

\section{Discussion and conclusion}

In scenarios where the training error diminishes to zero, such as in separable classes scenarios or with over-parameterized classifiers, numerous global minima with zero training error exist. We have shown that the proportion of "bad" global minima, which are characterized by a generalization error exceeding a certain threshold $\varepsilon$, decreases exponentially with the number of training samples $n$. The learning curve is closely related to the shape of the density of classifiers (DOC). The DOC depends on both the type of classification problem and the set of classifier functions used to solve the problem. The form of the DOC, in particular the normalized DOC, can be seen as the primary characteristic of a particular classifier/problem case. Increasing the absolute density values by expanding the capacity of the classifier function set can maintain the normalized DOC and, consequently, the learning curve, either invariant or even improve it.  This observation sheds light on the remarkable generalization capabilities of over-parameterized Deep Neural Networks, which seemed to be in contrast to generally accepted machine learning wisdom.

With some basic assumptions we developed a very general model of the DOC. This model provides learning curves with only two parameters, the minimal true error $E_{min}$ of the function set and a scaling factor $\eta$ that determines how quickly the generalization error converges to this minimal achievable value. $E_{min}$ can easily be reduced by increasing the capacity of the classifier. One would expect that this also increases $\eta$, with more training data necessary for convergence. However, in our experiments $\eta$ even decreased for ResNet. Increasing network size not only led to smaller errors, but also faster learning (with less data). This seemingly counterintuitive finding aligns with observations in \cite{Belkin2019}, where larger networks improved results in the over-parameterized regime. Intriguingly, this trend does not hold in our experiments with fully connected MLPs, possibly due to their shallower architecture compared to ResNets. Understanding how the classifier's structure, coupled with the classification problem, shapes the DOC is an intriguing avenue for future research.

In contrast to the "old tradition" in machine learning, our findings argue in favor of working consistently in the over-parameterized domain for three main reasons: (i) attaining a global minimum of the training error is typically straightforward, (ii) enlarging the network size readily reduces the minimal achievable generalization error, and (iii) when executed judiciously, this enhancement doesn't inherently demand additional training data to approach the optimal generalization error.

\section*{Acknowledgment}
The Version of Record of this contribution was presented at the 2024 International Joint Conference on Neural Networks (IJCNN),
and is available online at https://doi.org/10.1109/IJCNN60899.2024.10650680

\bibliographystyle{ieeetr}
\bibliography{references}

\newpage

\appendix
\section*{Appendix}

\subsection*{Arguments for approximation step}

Approximation step \eqref{eq:ProductAverages}
\begin{equation*}
    \left\langle \frac{\vert {\cal H}_\varepsilon({\cal S})\vert}{\vert {\cal H}({\cal S})\vert}\right\rangle_{{\cal S}} 
    \approx 
    \frac{\langle\vert {\cal H}_\varepsilon({\cal S})\vert\rangle_{{\cal S}}}{\langle\vert {\cal H}({\cal S})\vert\rangle_{{\cal S}}}
\end{equation*}
is a well-known first-order approximation based on the Taylor-expansion of $\frac{\vert {\cal H}_\varepsilon({\cal S})\vert}{\vert {\cal H}({\cal S})\vert}$ around $\langle\vert {\cal H}_\varepsilon({\cal S})\vert\rangle_{{\cal S}}$ and $\langle\vert {\cal H}({\cal S})\vert\rangle_{{\cal S}}$ (see, e.g., \cite{KendallStatistics2006}). This makes sense, if $\vert {\cal H}_\varepsilon({\cal S})\vert$ and $\vert {\cal H}({\cal S})\vert$ are concentrated around their means, which one expects at least for large $n$. 

Approximation step \eqref{eq:ProductAverages} is still valid, if only the ratio
\begin{equation}
    \frac{\vert {\cal H}_\varepsilon({\cal S})\vert}{\vert {\cal H}({\cal S})\vert} =f_{{\cal S}}(\varepsilon)
    \label{eq:ratio}
\end{equation}
is concentrated around its mean. The ratio $f_{{\cal S}}(\varepsilon)$ with $0\leq f_{{\cal S}}(\varepsilon) \leq 1$ quantifies the percentage of "bad" global minima for a given ${\cal S}$ and is a monotonically decreasing function of $\varepsilon$. Averaging over all ${\cal S}$ yields
\begin{equation}
    \left\langle \frac{\vert {\cal H}_\varepsilon({\cal S})\vert}{\vert {\cal H}({\cal S})\vert}\right\rangle_{{\cal S}} =f_n(\varepsilon),
    \label{eq:AverageRatio}
\end{equation}
which depends only on $n$. Equation~\eqref{eq:ratio} also yields
\begin{eqnarray}
    \left\langle \vert {\cal H}_\varepsilon({\cal S})\vert\right\rangle_{{\cal S}} &=&\left\langle f_{{\cal S}}(\varepsilon) \vert {\cal H}({\cal S})\vert\right\rangle_{{\cal S}} \nonumber \\
    &=&f_n(\varepsilon) \left\langle\vert {\cal H}({\cal S})\vert\right\rangle_{{\cal S}} + \hbox{cov}\left(f_{{\cal S}}(\varepsilon), \vert {\cal H}({\cal S})\vert\right)_{{\cal S}} \nonumber
\end{eqnarray}
with $\hbox{cov}(.,.)_{{\cal S}}$ as the covariance over all ${\cal S}$. But then, together with Equation~\eqref{eq:AverageRatio}, we obtain 
\begin{equation*}
    \frac{\langle\vert {\cal H}_\varepsilon({\cal S})\vert\rangle_{{\cal S}}}{\langle\vert {\cal H}({\cal S})\vert\rangle_{{\cal S}}}
    = 
    \left\langle \frac{\vert {\cal H}_\varepsilon({\cal S})\vert}{\vert {\cal H}({\cal S})\vert}\right\rangle_{{\cal S}} 
    +\frac{\hbox{cov}\left(f_{{\cal S}}(\varepsilon), \vert {\cal H}({\cal S})\vert\right)_{{\cal S}}}{\langle\vert {\cal H}({\cal S})\vert\rangle_{{\cal S}}}.
\end{equation*}
If for a given $\varepsilon$ the percentage $f_{{\cal S}}(\varepsilon)$ of "bad" global minima does not co-vary (too much) with the absolute size of the set of global minima $\vert {\cal H}({\cal S})\vert$, approximation step \eqref{eq:ProductAverages} holds. If the covariance is positive, i.e., if the percentage of "bad" global minima is larger for large $\vert {\cal H}({\cal S})\vert$, the r.h.s.\ of approximation step \eqref{eq:ProductAverages} is at least an upper bound. Since
\begin{equation*}
    \left\vert\hbox{cov}\left(f_{{\cal S}}(\varepsilon), \vert{\cal H}({\cal S})\vert\right)_{{\cal S}}\right\vert
        \leq \sqrt{\hbox{var}\left(f_{{\cal S}}(\varepsilon)\right)_{{\cal S}}}
        \;\sqrt{\hbox{var}\left(\vert{\cal H}({\cal S})\vert\right)_{{\cal S}}}\;,
\end{equation*}
approximation step \eqref{eq:ProductAverages} is also valid, if $f_{{\cal S}}(\varepsilon)$ is concentrated around its mean $f_n(\varepsilon)$ with a small variance. This is indeed observed in the experiments for large $n$. 

\subsection*{Exponential decay of Q(E)}

We show that $Q_n(E)$ decays to zero exponentially fast with increasing $n$ for each $E>E_{min}$. 

We have $D:\left[0,1\right]\rightarrow \mathbb{R}_0^+$, and $D(E)$ starts to become non-zero for $E\geq E_{min}$. According to Equation~\eqref{eq:Q_n} we have 
\begin{eqnarray*}
    Q_n(E)&=&\frac{(1-E)^n D(E)}{\int_{0}^1 (1-E')^n D(E')\,dE' }\\
        &=&\frac{D(E)}{\int_{0}^1 B(E')^n D(E')\,dE' }
\end{eqnarray*}
with 
\begin{equation*}
    B(E')=\frac{1-E'}{1-E}.
\end{equation*}
The case $E=1$ we can ignore, since then $Q_n(E)=0$ anyway. $B(E')$ decreases with increasing $E'$, and for $E'<E$, we have $B(E')>1$. Note, that always $0\leq E,E'\leq 1$ and, hence, $B(E')\geq 0$. 

Since $E_{min}<E$, there is an $a$ with $E_{min}<a< E$, and we obtain 
\begin{eqnarray*}
    Q_n(E)&=&\frac{D(E)}{\int_{0}^1 B(E')^n D(E')\,dE' }\\
        &\leq&\frac{D(E)}{\int_{0}^a B(E')^n D(E')\,dE' }\\
        &\leq&\frac{D(E)}{B(a)^n \int_{0}^a D(E')\,dE' }\\
        &\leq&\frac{D(E)}{\int_{0}^a D(E')\,dE' }B(a)^{-n}.
\end{eqnarray*}
Since $B(a)>1$, $Q_n(E)$ decays at least exponentially to zero with increasing $n$. 

\subsection*{Setup of experiments with over-parameterized classifiers on MNIST and CIFAR-10}
\label{sec:exp_setup}

\begin{table*}[tb]
    \caption{Detailed list of hyperparameters.}
	\label{tab:hyperparameters}
	\setlength\tabcolsep{2pt} 
	\begin{adjustbox}{width=1\linewidth}
    \begin{tabular}{@{}lllllllllllllllllll@{}}
    & & & & & & & & & & & & & & & & & &\\
                                     & n          & 10       & 20       & 50       & 100      & 200      & 500     & 1000    & 2000    & 5000 & 10000 & 15000 & 20000 & 25000 & 30000 & 40000 & 50000 & 60000 \\ \midrule
\multirow{3}{*}{\footnotesize{CIFAR-10, MLP3}}      & batch size & 10       & 20       & 50       & 100      & 100      & 100     & 100     & 100     & 1024 & 1024  & 1024  & 1024  & 1024  & 1024  & 1024  & 1024  & -     \\
                                     & epochs     & 1000     & 800      & 400      & 300      & 100      & 80      & 80      & 80      & 50   & 50    & 50    & 50    & 50    & 50    & 50    & 50    & -     \\ 
                                     & $1000 \cdot l_{\textrm{init}}$  & 0.3125   & 0.625    & 1.5625   & 3.125    & 3.125    & 3.125   & 3.125   & 3.125   & 16   & 16    & 16    & 16    & 16    & 16    & 16 & 16 & -     \\ \midrule
\multirow{3}{*}{\footnotesize{CIFAR-10, MLP8}}      & batch size & 10       & 20       & 50       & 100      & 100      & 100     & 128     & 128     & 128  & 128   & 128   & 128   & 128   & 128   & 128   & 128   & -     \\ 
                                     & epochs     & 1000     & 1000     & 500      & 400      & 100      & 100     & 80      & 80      & 50   & 50    & 50    & 50    & 50    & 50    & 50    & 50    & -     \\
                                     & $1000 \cdot l_{\textrm{init}}$  & 0.15625  & 0.3125   & 0.78125  & 1.5625   & 1.5625   & 1.5625  & 2       & 2       & 2    & 2     & 2     & 2     & 2     & 2     & 2 & 2 & -     \\ \midrule
\multirow{3}{*}{\footnotesize{CIFAR-10, ResNet018}} & batch size & 10       & 20       & 50       & 100      & 100      & 100     & 100     & 128     & 1024 & 1024  & 1024  & 1024  & 1024  & 1024  & 1024  & 1024  & -     \\
                                     & epochs     & 600      & 300      & 160      & 100      & 100      & 80      & 80      & 60      & 50   & 50    & 50    & 50    & 50    & 50    & 50    & 50    & -     \\ 
                                     & $1000 \cdot l_{\textrm{init}}$  & 0.15625  & 0.3125   & 0.78125  & 0.78125   & 0.78125  & 0.78125 & 0.78125 & 2       & 16   & 16    & 16    & 16    & 16    & 16    & 16 & 16 & -     \\ \midrule
\multirow{3}{*}{\footnotesize{CIFAR-10, ResNet101}} & batch size & 10       & 20       & 50       & 100      & 100      & 100     & 100     & 128     & 128  & 128   & 128   & 128   & 128   & 128   & 128   & 128   & -     \\
                                     & epochs     & 600      & 300      & 160      & 100      & 100      & 80      & 80      & 80      & 50   & 50    & 50    & 50    & 50    & 50    & 50    & 50    & -     \\
                                     & $1000 \cdot l_{\textrm{init}}$  & 0.15625  & 0.3125   & 0.78125  & 1.5625   & 1.5625  & 1.5625 & 0.15625 & 2       & 2    & 2     & 2     & 2     & 2     & 2     & 2 & 2 & -     \\ \midrule 
\multirow{3}{*}{\footnotesize{MNIST, ResNet018}}    & batch size & 10       & 20       & 50       & 100      & 100      & 100     & 100     & 100     & 1024 & 1024  & 1024  & 1024  & 1024  & 1024  & 1024  & 1024  & 1024  \\
                                     & epochs     & 1000     & 600      & 300      & 100      & 100      & 80      & 80      & 80      & 50   & 50    & 50    & 50    & 50    & 50    & 50    & 50    & 50    \\
                                     & $1000 \cdot l_{\textrm{init}}$  & 0.078125 & 0.15625  & 0.390625 & 0.78125  & 0.78125  & 0.78125 & 0.78125 & 0.78125 & 8    & 8     & 8     & 8     & 8     & 8     & 8 & 8 & 8 \\ \midrule
\multirow{3}{*}{\footnotesize{MNIST, ResNet101}}    & batch size & 10       & 10       & 10       & 10       & 10       & 100     & 128     & 128     & 128  & 128   & 128   & 128   & 128   & 128   & 128   & 128   & 128   \\
                                     & epochs     & 600      & 500      & 500      & 500      & 500      & 100     & 100     & 80      & 50   & 50    & 50    & 50    & 50    & 50    & 50    & 50    & 50    \\
                                     & $1000 \cdot l_{\textrm{init}}$  & 0.078125 & 0.078125 & 0.078125 & 0.078125 & 0.078125 & 0.78125 & 1       & 1       & 1    & 1     & 1     & 1     & 1     & 1     & 1 & 1 & 1
    \end{tabular}
    \end{adjustbox}
    \setlength\tabcolsep{6pt} 
\end{table*}

\begin{table}[tb]
	\caption{ResNet architectures. Feature map sizes are shown for input images with $32 \times 32$ pixels.}

	\label{tab:resnet_architectures}
	\centering
		\begin{tabular}{ccm{3cm}m{3cm}}
		    & & & \\
			\textbf{Layers} & \textbf{Output size} & \textbf{ResNet018} & \textbf{ResNet101} \\
			\midrule
			
			Convolution & $32 \times 32$ &  $3 \times 3$, 64, stride 1 &  $3 \times 3$, 64, stride 1   \\
			
			\midrule
			\multirow{3}{*}{ResNet Block (1)} & 
			\multirow{3}{*}{$32 \times 32$} &  
			\multirow{3}{*}{
				$\begin{bmatrix}
				3 \times 3, 64 \\
				3 \times 3, 64
				\end{bmatrix} \times 2$ } &
			\multirow{3}{*}{
				$\begin{bmatrix}
				1 \times 1, 64\\
				3 \times 3, 64 \\
				1 \times 1, 256
				\end{bmatrix} \times 3$ } \\
				& & \\
				& &	\\
			
			\midrule
			\multirow{3}{*}{ResNet Block (2)} & \multirow{3}{*}{$16 \times 16$} &  
			\multirow{3}{*}{
				$\begin{bmatrix}
				3 \times 3, 128\\
				3 \times 3, 128
				\end{bmatrix} \times 2$ } &
			\multirow{3}{*}{
				$\begin{bmatrix}
				1 \times 1, 128\\
				3 \times 3, 128 \\
				1 \times 1, 512
				\end{bmatrix} \times 4$ } \\
				& & \\
				& &	\\

			\midrule
			\multirow{3}{*}{ResNet Block (3)} & \multirow{3}{*}{$8 \times 8$} &  
			\multirow{3}{*}{
				$\begin{bmatrix}
				3 \times 3, 256\\
				3 \times 3, 256
				\end{bmatrix} \times 2$ } &
			\multirow{3}{*}{
				$\begin{bmatrix}
				1 \times 1, 256\\
				3 \times 3, 256 \\
				1 \times 1, 1024
				\end{bmatrix} \times 23$ } \\
				& & \\
				& &	\\

			\midrule
			\multirow{3}{*}{ResNet Block (4)} & \multirow{3}{*}{$4 \times 4$} &  
			\multirow{3}{*}{
				$\begin{bmatrix}
				3 \times 3, 512\\
				3 \times 3, 512
				\end{bmatrix} \times 2$ } &
			\multirow{3}{*}{
				$\begin{bmatrix}
				1 \times 1, 512\\
				3 \times 3, 512 \\
				1 \times 1, 2048
				\end{bmatrix} \times 3 $ } \\
				& & \\
				& &	\\

			\midrule		
			\multirow{2}{*}{Classification layer} & $1 \times 1$ &  $4\times 4$ Adaptive avg. pool & $4\times 4$ Adaptive avg. pool \\
			\cmidrule(l){2-4}
							& 		& 	fully connected, softmax	& 	fully connected, softmax \\
		\end{tabular}
\end{table}

The experiments on separable classes are straightforward to implement. The experiments with the over-parameterized classifiers are more complex. In the following we provide all the details for reimplementation.

The ResNet018 and ResNet101 architectures are slightly adjusted to improve the processing of small-sized images. We do not place a max-pooling layer between the first convolutional layer and the first residual block, and we change the kernel size of the first convolutional layer to $3 \times 3$ and apply a stride of 1. Details about the architectures are provided in Table \ref{tab:resnet_architectures}. For the convolutional layers we use the Kaiming initialization \cite{he2015delving} with uniform distribution. It is worth mentioning that the images from the CIFAR-10 dataset are of shape $32 \times 32 \times 3$, while MNIST contains smaller gray-scale images with $28 \times 28 \times 1$ pixels. In order to use the ResNet architectures on both datasets, we convert the gray-scale images to RGB with $R=G=B$. Thus, the ResNet architectures for MNIST and CIFAR-10 are identical.

The smaller MLP3 network begins with two fully connected layers of $28 \times 28 \times 3$ neurons each. The third layer has 10 output neurons, reflecting the number of classes. Between each layer, the ReLU activation function \cite{nair2010rectified} is applied. The larger MLP8 model differs from its smaller counterpart only in the number of hidden layers, which is seven instead of two.

Image normalization is performed by subtracting the average color value of the CIFAR-10/MNIST dataset and dividing by the standard deviation in a channel-wise manner. We do not perform any augmentation, because we explicitly do not want to artificially enlarge the training dataset.

For the purpose of statistical evaluation, we train 50 networks per combination of dataset, architecture, and number of training samples $n\in$ \{10, 20, 50, 100, 200, 500, 1000, 2000, 5000, 10000, 15000, 20000, 25000, 30000, 40000, 50000, (60000)\}. The 50 networks differ in training samples and initial weights. The n training samples are randomly drawn for each network, such that each class is represented by the same number of images. We test on the entire test dataset, each of which contains 10000 images. This means that for $n=20$ we train on two training images per class and test on all 10000 test images. 

We train the models using the Pytorch framework \cite{NEURIPS2019_9015}, cross-entropy loss and the LAMB optimizer \cite{you2019large} with parameters $\beta_1 = 0.9$ and $\beta_2 = 0.999$ without any weight decay. After training, all networks classify their training samples correctly. The learning rate is initialized with $l_{\textrm{init}}$ and decays to a minimum of $l_{\textrm{init}} / 1000$. We used large batch sizes to keep the training time feasible, although, since the batch size can at most be as large as $n$, we had to decrease the batch size for smaller $n$. At the same time it was necessary to increase the number of epochs. A detailed list of hyperparameters can be found in Table~\ref{tab:hyperparameters}. 

\end{document}